\documentclass[letterpaper, 10 pt, conference]{ieeeconf}
\IEEEoverridecommandlockouts    
\usepackage{graphics}           
\usepackage{times}              
\usepackage{amsmath}            
\usepackage{amssymb}            
\usepackage{graphicx}
\usepackage{svg}
\usepackage{algorithm}
\usepackage[noend]{algpseudocode}
\usepackage{booktabs}
\usepackage{color}
\usepackage[nocompress]{cite} 
\definecolor{instructioncolor}{rgb}{.5,.5,.5}

\usepackage[font=small]{caption}
\usepackage{hyperref}

\def\tabref#1{Tab.~\ref{#1}}
\def\eqref#1{Eq.~(\ref{#1})}

\makeatletter
\usepackage{xspace}
\DeclareRobustCommand\onedot{\futurelet\@let@token\@onedot}
\def\@onedot{\ifx\@let@token.\else.\null\fi\xspace}

\def\etal{{et al}\onedot}
\makeatother

\usepackage{array}
\newcolumntype{L}[1]{>{\raggedright\let\newline\\\arraybackslash\hspace{0pt}}m{#1}}
\newcolumntype{C}[1]{>{\centering\let\newline\\\arraybackslash\hspace{0pt}}m{#1}}
\newcolumntype{R}[1]{>{\raggedleft\let\newline\\\arraybackslash\hspace{0pt}}m{#1}}

\usepackage{float} 
\usepackage{multirow} 

\title{\LARGE \bf Crop Spirals: Re-thinking the field layout for future robotic agriculture}

\author{Lakshan L.$^{1}$ \and Lanojithan T.$^{1}$\and Udara Muthugala$^{1}$ \and Rajitha de Silva $^{2}$
  \thanks{$^{1}$Lakshan L., Lanojithan T. and Udara Muthugala are with Faculty of Technology, University of Colombo, Sri Lanka. $^{2}$Rajitha de Silva is with Lincoln Centre for Autonomous Systems (L-CAS), University of Lincoln, UK. (correspondence author: Rajitha de Silva)
        {\tt\small $^{4}$odesilva@lincoln.ac.uk}
  }%
}
\begin{document}
\maketitle
\thispagestyle{empty}
\pagestyle{empty}

\begin{abstract}
  %



  

Conventional linear crop layouts, optimized for tractors, hinder robotic navigation with tight turns, long travel distances, and perceptual aliasing. We propose a robot-centric \emph{square spiral} layout with a central tramline, enabling simpler motion and more efficient coverage. To exploit this geometry, we develop a navigation stack combining DH-ResNet18 waypoint regression, pixel-to-odometry mapping, A* planning, and model predictive control (MPC). In simulations, the spiral layout yields up to \textbf{28\%} shorter paths and about \textbf{25\%} faster execution for waypoint-based tasks across 500 waypoints than linear layouts, while full-field coverage performance is comparable to an optimized linear U-turn strategy. Multi-robot studies demonstrate efficient coordination on the spiral’s rule-constrained graph, with a greedy allocator achieving \textbf{33--37\%} lower batch completion times than a Hungarian assignment under our setup. These results highlight the potential of redesigning field geometry to better suit autonomous agriculture.
\end{abstract}

\section{Introduction}
\label{sec:intro}


Since the invention of the steam engine, tractors were developed and gradually introduced into farming. Their arrival profoundly revolutionized agricultural practices, especially in arable farming~\cite{ref_tractor_history}. This mechanization substantially increased throughput and efficiency, shaping the conventional layout that featured parallel rows optimized for tractor movement~\cite{ref_spekken}. However, decades of heavy machinery use have degraded soils through compaction, biodiversity loss, and reduced water infiltration~\cite{ref_botta2022, ref_parfitt2014, ref_shaheb2021}. With the growing global trend towards agricultural robotics~\cite{ref_raj2024, ref_albiero2020}, there is an opportunity to reduce dependence on heavy machinery by deploying lightweight autonomous robots~\cite{ref_mansur2025}. While a single robot cannot match the throughput of a tractor, coordinated multi-robot systems offer scalable alternatives~\cite{ref_aguiar2022}. Therefore, rethinking field layout for crops to accommodate multi-robot systems is essential for sustainable and efficient robotic agriculture, addressing these environmental concerns and enhancing operational efficiency.

\begin{figure}[t] \centering \includegraphics[width=\linewidth]{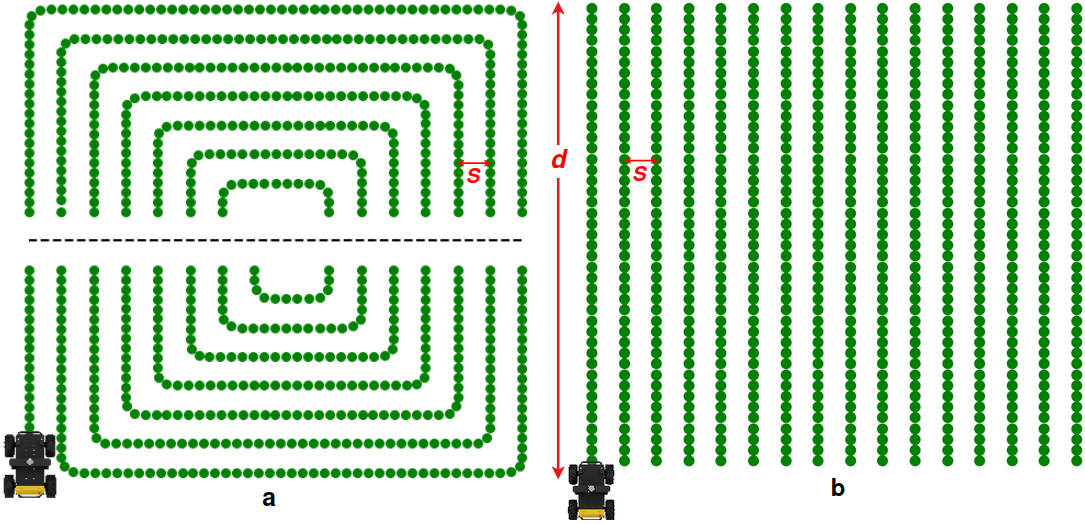} \caption{Field layout comparison. (a) Proposed square-spiral layout with central tramline(\texttt{---}) and (b) the conventional linear field layout. \textit{s}:Inter-row separation, \textit{d}:Width of the square field.} \label{fig:spiral_vs_linear} \end{figure}


The traditional ``linear field layout", as referred to in this paper, is characterised by fieldwork tracks, which are parallel crop rows aligned with the longitudinal axis of the field. In some cases, this layout also incorporates headland passes running parallel to the field boundary, which facilitate tractor turning, transitions between fieldwork tracks, and field entry or exit~\cite{vahdanjoo2020route}. Despite their historical success, linear layouts optimised for tractors pose challenges in perception and navigation of modern autonomous robots. At the end of rows, robots must perform tight turns, which are difficult to perform reliably and often lead to inefficiencies or crop damage in the headland pass areas~\cite{ref_peng2024, ahmadi2020visual}. Furthermore, the highly repetitive geometry of crop rows causes perceptual aliasing, where visual sensors encounter nearly identical views along the row, making it difficult to distinguish the precise location of the robot~\cite{ref_aguiar2020}. This problem is further amplified at the end of the row, where the forward facing camera setups no longer detect crops, rendering row following methods ineffective and forcing reliance on alternative navigation cues such as odometry, Global Positioning System (GPS) triggers, or predefined turn-by-turn waypoints. Linear layouts also concentrate inter-row transfers, causing bottlenecks, dead-head travel, and inefficient multi-robot coordination~\cite{ref_bochtis2008,ref_tu2019,ref_wang2023,ref_hoffmann2023}. To this end, adapting agricultural practices for robotic technology requires fundamental revisions of traditional crop field layouts.


While existing work typically concentrate on improving perception and navigation algorithms for repetitive linear field layout, no studies have directly targeted the underlying field geometry. Historically, field layouts have evolved to match prevailing technologies; we propose a square spiral layout that incorporates a tramline, enabling efficient access to any field coordinate with significantly reduced travel distances and minimal navigational complexity as illustrated in Fig.~\ref{fig:spiral_vs_linear}. This layout also simplifies the two-dimensional (2D) navigation problem in field environments to a 1D navigation problem, as a 2D spiral could be projected on to a 1D coordinate space. This also injects sufficient entropy into the field structure enabling previously constrained navigation methods such as teach and repeat schemes~\cite{cox2023visual} and 1D Monte-Carlo localisation to be used in agri-robotic navigation.



The main contribution of this paper is the introduction of a novel square spiral crop layout as an alternative to conventional linear field layout, uniquely optimized for robot-friendly navigation. Specifically, we propose: (i) a vision-based navigation path prediction system uniquely suited for spiral fields, (ii) a path planning algorithm for proposed spiral layout, and (iii) a novel decentralized multi-robot coordination framework designed explicitly for scalable operation within spiral layouts.

\noindent
In summary, we make three experimentally supported claims.
(i) In waypoint-based tasks the square-spiral layout cuts mean travel distance by
$\approx28\%$ and travel time by $\approx25\%$ relative
to a linear field, implying lower mechanical energy use.
(ii) Our vision-MPC stack delivers higher accuracy, achieving 0.38m mean absolute position error in full-field runs and a 50\% improvement on heading error with the DH-ResNet18.
(iii) The layout and control framework scales to multi-robots: in our tests a decentralized \emph{Greedy} allocator cut batch completion times by $33$--$37\%$ compared with the Hungarian method while keeping slightly lower workload-fairness CV, and the biggest throughput gain came when scaling from two to three robots.

\section{Related Work}
\label{sec:related}


The origins of modern field layout are widely traced to English agriculturalist Jethro Tull, whose invention of the horse-drawn seed drill is considered a pivotal breakthrough in agricultural mechanisation~\cite{sayre2010pre}. His systematic planting of seeds in evenly spaced rows fundamentally formalised the principles of modern crop field layout. These linear field layouts, eventually optimized for tractor-based cultivation with uniform rows and headland turns. This modern field layout optimized for tractor models poses critical challenges for autonomous robot navigation. Long uninterrupted rows increase travel distances for smaller robots and narrow headlands complicate turning between rows~\cite{ref_baby2024,ref_sivakumar2021}. Researchers have explored contour farming, strip cropping, and headland optimization~\cite{ref_donat2023,ref_mier2025,ref_hoffmann2023}, and a few studies attempt layout reconfiguration to ease robotic motion~\cite{ref_schmitz2022}, but these approaches remain rooted in tractor-first paradigms. Our work builds on this gap by proposing a spiral geometry that minimizes tight turns and navigation inefficiencies while enabling efficient, robot-friendly movement.

Vision-based perception has been a key enabler in autonomous agricultural robotics~\cite{bai2023vision}. However, the perceptual aliasing problem in robotic perception is a common hurdle highlighted by many researchers due to the repetitive nature of the linear field structure~\cite{de2024deep, ahmadi2020visual}. Deep learning methods using red–green–blue (RGB) or multispectral imagery demonstrate robust performance in plant detection, row following, and crop–weed classification~\cite{zhang2024review,de2024vision,bakker2008vision, fawakherji2021multi}. Convolutional neural network (CNN) architectures, such as ResNet~\cite{he2016deep}, are now widely used for perception tasks ranging from semantic segmentation to navigation path detection~\cite{wang2018understanding, sivakumar2021learned}. Yet most prior vision systems are tuned for structured environments with pre-programmed row layouts~\cite{bonadies2019overview, shi2023row}, limiting their adaptability to alternative geometries like spirals. Most existing vision-based systems follow reactive navigation rather than classical localization, mainly due to the repetitive nature of linear field layouts~\cite{bai2023vision, bonadies2019overview}. Our method extends this body of work by predicting ego trajectory directly from camera images, specifically designed for spiral-field layout.

Following a predicted ego-trajectory can be accomplished using a range of controllers, from classical proportional–integral–derivative (PID) schemes to more advanced formulations such as MPC. Spiral coverage paths, which alternate between straight passes and sharp 90\textdegree~
turns, place varying demands on the controller; smooth tracking therefore requires adaptation to changing motion profiles. MPC is well suited to this setting, as it explicitly accounts for system dynamics and constraints when following time-varying trajectories~\cite{lu2024path}. Demonstrated applications include vineyard and orchard operations~\cite{vatavuk2022task, wang5288483field}, while only a limited number of studies have investigated curvature-constrained planners, for agricultural navigation~\cite{he2023dynamic}. Recent work in coverage path planning has emphasised minimising overlap and energy consumption~\cite{fevgas2022coverage}, but these approaches often neglect the role of field geometry. Our spiral-aware planner addresses this gap by incorporating constraints optimised for the spiral geometry within a predictive control framework, thereby improving both path smoothness and coverage efficiency.

The transition from tractor-based operations to autonomous robotics in agriculture is expected to rely on multi-robot systems rather than a single small robot covering an entire field. Such systems enable scalability and reduce labour demands, and have consequently attracted growing research interest~\cite{mao2021research}. Prior frameworks explore decentralized coordination, consensus-based strategies, and market-driven task allocation~\cite{10333601,roberts2025kriging}. More recent studies adopt partition-based coordination, in which the field is divided into subregions and robots rely on local dead-reckoning within their assigned partitions~\cite{choton2024coverage}; however, these methods are typically designed for conventional linear field layout. In contrast, our approach leverages the segmented structure of the square-spiral layout, comparing a greedy decentralized strategy with a centralized Hungarian assignment~\cite{kuhn1955hungarian} to achieve efficient, collision-free cooperation.

Finally, robotics for sustainable agriculture addresses the environmental costs of heavy machinery. Lightweight autonomous platforms reduce soil compaction and fuel emissions while enhancing soil biodiversity~\cite{bruvciene2025comparative,ghobadpour2022off}. Complementary research in robotic weeding, precision fertilization, and minimal tillage further illustrates how automation can align with ecological goals~\cite{sosnoskie2025deep,munnaf2024robot}. Beyond these environmental gains, autonomy also changes the economics of field operations. Recent analyses show that autonomous crop machines perform more efficiently in smaller, irregularly shaped fields than conventional tractors, challenging the historical trend toward land consolidation~\cite{al2023economics}. This shift suggests that robotics not only supports sustainable practices but also redefines the economic viability of diverse field geometries. Our approach contributes at an upstream level by redesigning the field layout, enabling both environmental and operational improvements in robot-based agriculture.

\begin{figure*}[t]
  \centering
  \includegraphics[width=\textwidth]{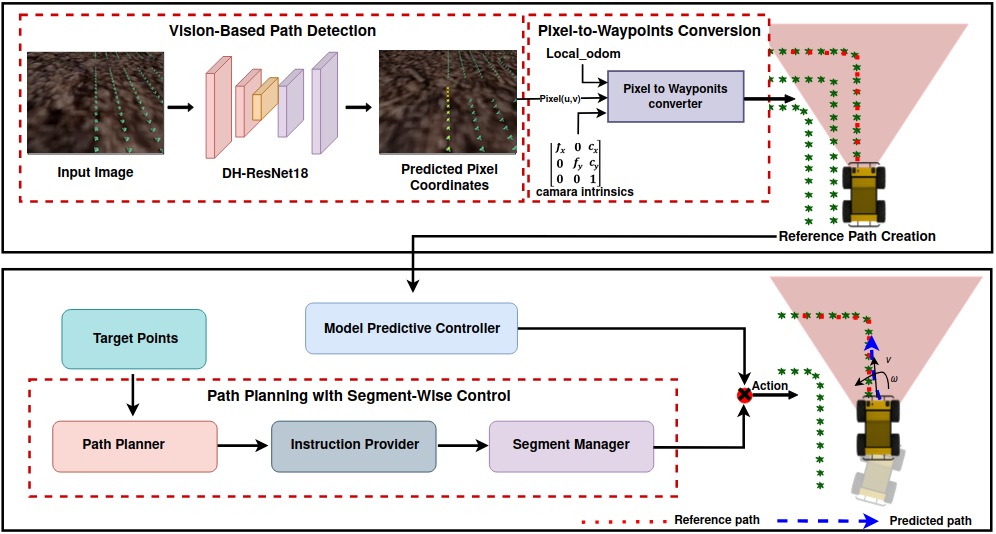}
  \caption{Overview of the proposed navigation system: integration of DH-ResNet18–based waypoint prediction, pixel-to-odometry transformation, path planning,
  segment-wise control, and MPC-based trajectory tracking.}
  \label{fig:system_overview}
\end{figure*}


\section{Navigation in Spiral Crop Fields}
\label{sec:main}

We propose a square-spiral field layout as a new paradigm to support efficient navigation and enhanced perception in agricultural robotics. Building on this layout, we design an integrated navigation framework that improves crop accessibility and operational efficiency.  The system combines a multi-task ResNet-based perception module, which simultaneously predicts ego-trajectory waypoints and classifies scene context from raw RGB images, where scene classification distinguishes between spiral straight segments and spiral bend regions, with a path planning pipeline tailored to the spiral’s constrained topology. Since visiting multiple crop segments of the spiral can be formulated as a Traveling Salesman Problem (TSP), we employ A* search for local path generation and a brute-force TSP solver for globally optimal multi-target routing. Trajectory execution is handled by an MPC controller, which incorporates real-time feedback and segment-specific control logic to maintain accuracy. Collectively, these components form a modular, adaptable framework tailored to structured agricultural environments, and demonstrates clear potential for scalable multi-robot collaboration. An overview of the architecture is shown in Fig.~\ref{fig:system_overview}.


\subsection{Spiral Field Layout}

We propose a spiral field layout in which the crop row begins at a field corner and unfolds in a rectangular spiral with rounded corners, continuously covering the entire area while maintaining equal spacing between successive passes (Fig.~\ref{fig:spiral_vs_linear}.a). At the centre of the spiral, a tramline provides a direct-access path that allows robots to reach any location in the field efficiently, enabling rapid interventions such as targeted spraying or point-specific agronomic tasks. Although the concept can be generalised to arbitrary field shapes under the same rules, in this work we present an initial case study using a square plot for simplicity. Accordingly, we refer to this configuration as the “square-spiral” layout, representing the simplified version explored in this paper.

Compared to conventional linear field layout with headland turns, the proposed spiral layout increases effective crop density by eliminating the dedicated turn zones where tractors typically execute multiple manoeuvres, often causing crop damage. The geometry of the square-spiral \(\mathcal{S}(s,d)\) is determined by the side length of the square plot 
\( d \), and the inter-row spacing, 
\( s \). These parameters define the number of crop row segments as

\begin{equation}
n = \left\lfloor \frac{d}{s} \right\rfloor + 1
\label{eq:num_rows}
\end{equation}

which corresponds to a total of \(2n - 1\) straight row segments in the layout. We provide an open-source toolkit\footnote{\url{https://github.com/rajithadesilva/spiral_crops}} that supports defining and visualising square spirals, performing 2D-to-1D coordinate transformations, calculating lengths, characterising individual segments and simulating the proposed pipeline in this paper.

\subsection{Vision-Based Navigation in a Spiral Field}
\label{sec:vision}
A multi-task CNN, termed Dual-Head ResNet-18 (DH-ResNet18), was developed for real-time waypoint prediction and scene classification from monocular RGB images. We adopt a ResNet-18 backbone pretrained on ImageNet, removing its global-average-pooling and final classification layers. To the last convolutional feature map, we add a $3\times3$ convolution (512$\rightarrow$64 channels, stride~2) with batch normalization and ReLU activation, followed by a shared 128-unit fully connected layer and a lightweight dual output head that comprises (i) a regression branch that outputs a 20-dimensional vector representing ten \((u,v)\) pixel coordinate pairs and (ii) a classification branch that predicts the type of scene: straight segment, left bend or right bend.

This could be modelled as:
\begin{equation}
\hat{\mathbf{y}}, \hat{c} = f_{\theta}(I),
\end{equation}
where $f_{\theta}$ is the CNN with parameters~$\theta$, $I \in \mathbb{R}^{3 \times 240 \times 320}$ is the input image, $\hat{\mathbf{y}} \in \mathbb{R}^{20}$ are the regressed pixel coordinates (down-scaled from $640 \times 480$ annotations and re-scaled for deployment), and $\hat{c} \in \mathbb{R}^{3}$ are the unnormalized class logits.

Training minimizes the sum of a Smooth-L1 regression loss and a cross-entropy classification loss:
\begin{equation}
\mathcal{L}_{\text{total}}
  = \mathcal{L}_{\text{reg}}^{\text{Smooth-L1}}
  + \mathcal{L}_{\text{cls}}^{\text{CrossEntropy}} .
\end{equation}
Input images are augmented with random $\pm5^{\circ}$ rotations, color jitter, and ImageNet mean–std normalization. The network is trained with the Adam optimizer (initial learning rate $1\times10^{-3}$, weight decay $1\times10^{-5}$) and a cosine-annealing learning-rate schedule over 100 epochs.

Although the CNN outputs image-space predictions, a projection module converts the pixel coordinates back to the odometry frame for downstream trajectory tracking by the MPC controller (Section~\ref{mpc}), enabling robust, real-time perception-driven navigation in the proposed spiral field.

\subsection{Path Planning in Spiral Layout}
\label{sec:path_planning}
The spiral field is modeled as a weighted graph \(G=(V,E)\), where the
vertices \(V\) are discretized \emph{2-D waypoints} (planar coordinates)
along the 15 spiral navigation lanes and the central tramline. Edges \(E\) connect
successive vertices, each weighted by Euclidean distance
\[
w(a,b)=\lVert a-b\rVert_{2}.
\]

Each lane permits traversal \emph{only} from its entry vertex to its exit vertex in the forward direction.
Each lane’s designated \emph{entry} and \emph{exit} waypoints are fixed during map
construction, so their roles are unambiguous.  Inter-lane transfers are permitted
\emph{only} through the tramline, ensuring rule-based and directionally consistent
traversal. Any continuous target coordinates are first snapped to the nearest
vertex in \(V\).

Shortest paths are computed with A* search using the admissible and consistent
Euclidean heuristic, which guarantees optimality:
\begin{equation}
f(n) = g(n) + h(n),
\label{eq:astar_cost}
\end{equation}
where \(g(n)\) is the accumulated path cost from the start to \(n\), and \(h(n)\)
is the straight-line distance from \(n\) to the goal.

For multi-target planning, let the robot start at vertex \(s\) and let
the target set be \(T=\{t_1,\dots,t_k\}\), where each lowercase \(t_i\) denotes a
single target vertex. A brute-force strategy evaluates all
permutations \(\pi\) of \(T\) and selects the order minimizing the total constrained
tour cost:
\begin{equation}
\pi^{*} = \arg\min_{\pi}
\Big[ d_G^{\mathrm{rules}}\!\big(s,\pi(1)\big) +
\sum_{i=1}^{k-1} d_G^{\mathrm{rules}}\!\big(\pi(i),\pi(i+1)\big) \Big],
\label{eq:tsp_opt}
\end{equation}
where \(d_G^{\mathrm{rules}}(u,v)\) denotes the constrained shortest-path distance
between vertices \(u,v\in V\).  This exhaustive search is computationally feasible
because each planning batch contains only a small number of targets.  In other
words, \(d_G^{\mathrm{rules}}\) is the path length returned by A* when one-way lane
directions and tramline-only inter-lane transfers are enforced equivalently,
the length of the shortest valid route the robot can traverse without violating
these rules.

\subsection{Path Tracking Model Predictive Controller}
\label{mpc}
To navigate the structured spiral layout with high precision, a MPC strategy is employed. The controller predicts future robot states and computes optimized control commands over a finite horizon, adapting to curvature and motion constraints in real time.

The robot’s state vector at timestep \(k\) is defined as
\begin{equation}
\mathbf{x}_k = [\,x_k,\, y_k,\, \theta_k\,]^\top,
\end{equation}
and the control input vector as
\begin{equation}
\mathbf{u}_k = [\,v_k,\, \omega_k\,]^\top,
\label{eq:mpc_state_input}
\end{equation}
where \((x_k, y_k)\) denote the Cartesian position, \(\theta_k\) the heading, and \(v_k,\omega_k\) the linear and angular velocity commands.

The MPC minimizes a quadratic cost function over a prediction horizon \(N\), penalizing tracking error, control effort, and terminal deviation:
\begin{equation}
J = \sum_{k=0}^{N-1} \left( \mathbf{e}_k^\top Q \mathbf{e}_k + \mathbf{u}_k^\top R \mathbf{u}_k \right) + \mathbf{e}_N^\top Q_f \mathbf{e}_N,
\label{eq:mpc_cost_terminal}
\end{equation}
where \(\mathbf{e}_k = \mathbf{x}_k - \mathbf{x}_k^{\text{ref}}\) denotes the pose error, and \(Q, R, Q_f\) are positive-definite weighting matrices. For brevity, additional penalty terms used in the implementation, namely reference velocity tracking and control rate smoothing, are omitted from ~\eqref{eq:mpc_cost_terminal}.

This MPC formulation enables smooth, real-time trajectory tracking by closing the loop on robot state feedback while following reference paths predicted by the DH-ResNet18 based visual perception module (see Section~\ref{sec:vision}).

\subsection{Multi-Robot Collaboration for Spiral Fields}
\label{sec:multi_robot}
Building on the \emph{rule-constrained graph} \(G=(V,E)\) and the
\emph{distance metric} \(d_G^{\mathrm{rules}}(\cdot,\cdot)\)
introduced in Section~\ref{sec:path_planning}, we design two complementary
strategies for coordinating multiple robots within the spiral crop field.
Recall that \(G\) encodes all one-way lane directions and the tramline-only
transfer rules, and that \(d_G^{\mathrm{rules}}(u,v)\) is the
shortest valid path length returned by the A* search of ~\eqref{eq:astar_cost} while enforcing those constraints.

\subsubsection{Greedy Decentralised}
\label{sec:greedy_decentralised}
Each robot \(R_i\), with planar pose \(p_i(t)\in\mathbb{R}^2\) at time step \(t\),
selects its next unvisited target from the current free set
\(\mathcal{T}_{\mathrm{free}}\) according to
\begin{equation}
T_i^{*}
  = \arg\min_{T_j\in\mathcal{T}_{\mathrm{free}}}
      d_G^{\mathrm{rules}}\!\bigl(p_i(t),T_j\bigr),
\label{eq:greedy}
\end{equation}
where the rule-constrained distance \(d_G^{\mathrm{rules}}\) is exactly the
A*-based metric defined in Section~\ref{sec:path_planning}.
The set \(\mathcal{T}_{\mathrm{free}}\) contains all targets not yet reserved by
other robots.  Lanes currently being serviced are marked \emph{locked}, and any
target already claimed is \emph{reserved} and removed from the free set.

To avoid collisions on the central tramline, the next-step waypoints of every
robot pair are predicted. If both robots are on, or about to enter, the
tramline and their predicted centre–to–centre distance satisfies the safety
condition below, the robot with the longer remaining route yields for a fixed
pause \(W\); ties are broken deterministically by robot identifier. Here
\(p_i(t{+}1)\) denotes the predicted centre of robot \(i\) one step ahead,
\(r_s\) is the safety radius of a single robot, and the factor \(2 r_s\)
represents the combined clearance needed for two robots:
\begin{equation}
\bigl\| p_i(t{+}1) - p_j(t{+}1) \bigr\|_{2} < 2\,r_s
\;\Rightarrow\;
\begin{aligned}[t]
 &\text{longer-path robot}\\[-1ex]
 &\text{waits for \(W\) time units}.
\end{aligned}
\label{eq:collision}
\end{equation}

\subsubsection{Hungarian Centralised}
\label{sec:hungarian_centralised}
Alternatively, a central allocator assigns targets in discrete batches by solving
\begin{equation}
\begin{aligned}
\min_{x_{ij}\in\{0,1\}}\;& \sum_{i}\sum_{j} C_{ij}\,x_{ij}\\
\text{s.t. }& \sum_{j} x_{ij}\le 1,\;
             \sum_{i} x_{ij}\le 1 ,
\end{aligned}
\label{eq:hungarian}
\end{equation}
where \(x_{ij}=1\) if target \(T_j\) is assigned to robot \(R_i\).
For each lane-level task the assignment cost is
\[
C_{ij}=\min\bigl(
d_G^{\mathrm{rules}}(p_i,t_{j,\text{entry}}),
d_G^{\mathrm{rules}}(p_i,t_{j,\text{exit}})
\bigr),
\]
again using the rule-constrained distance of
Section~\ref{sec:path_planning}; infeasible pairs are masked by setting
\(C_{ij}=+\infty\).
After the batch allocation, \emph{each robot independently sequences its
assigned targets using a nearest-neighbour heuristic that respects the
lane-direction and tramline constraints of Section~\ref{sec:path_planning}},
ensuring efficient intra-batch traversal while keeping the global allocation
optimal.

\paragraph*{\textbf{Shared Collision Layer}} 
\label{sec:shared_collision}
Both strategies run atop the common collision-avoidance layer of ~\eqref{eq:collision}, ensuring safe and cooperative yielding on the
central tramline.

\paragraph*{\textbf{Performance Metrics}}
Let \(m\) denote the number of robots.
To compare the Greedy and Hungarian strategies, we evaluate two indicators,
both to be minimised:

\begin{itemize}
  \item \textbf{Average batch completion time:}  
        Let \(B\) be the total number of task batches and \(BT_b\) the completion
        time of batch \(b\) (wall-clock time from batch start until the last
        robot finishes its assigned targets).  
        The overall time metric is
        \begin{equation}
        \overline{BT} = \frac{1}{B}\sum_{b=1}^{B} BT_b .
        \label{eq:avg_batch_time}
        \end{equation}

  \item \textbf{Workload fairness:}  
        For each batch \(b\), let \(Z_{b,r}\) be the integer count of targets
        served by robot \(r\) out of \(m\) robots. Define
        \[
        \mu_b = \frac{1}{m}\sum_{r=1}^{m} Z_{b,r}, \qquad
        \sigma_b = \sqrt{\frac{1}{m}\sum_{r=1}^{m}\bigl(Z_{b,r}-\mu_b\bigr)^2}.
        \]
        The coefficient of variation for batch \(b\) is
        \begin{equation}
        \mathrm{CV}_b = \frac{\sigma_b}{\mu_b},
        \end{equation}
        and the experiment-level fairness score is the mean
        \begin{equation}
        \overline{\mathrm{CV}}
          = \frac{1}{|\mathcal B|}\sum_{b\in\mathcal B} \mathrm{CV}_b ,
        \label{eq:workload_cv}
        \end{equation}
        where \(\mathcal B = \{ b : \mu_b > 0 \}\) is the set of batches
        in which at least one target was allocated.
\end{itemize}

\section{Experimental Evaluation}
\label{sec:exp}

\subsection{Experimental Setup}

The experimental evaluation was conducted in two simulation setups to validate the proposed navigation framework. The Gazebo simulator (Fig.~\ref{fig:exp_envs}.a) was employed for testing perception and navigation (Experiments 1 \& 2), while a lightweight 2D simulator (Fig.~\ref{fig:exp_envs}.b) was used for the multi-robot navigation experiment (Experiment 3).

\subsubsection{Gazebo Spiral and Linear Field Navigation}

The first environment was the Gazebo simulator, where the standard Clearpath Husky A200 robot model was deployed in two crop layouts: a conventional linear field and the proposed spiral field. As shown in Fig.~\ref{fig:exp_envs}.a, the spiral layout followed configuration \(\mathcal{S}(0.75,11.5)\), while the linear layout was configured to match the same field size and row spacing of 0.75m, resulting in 16 parallel crop rows, each 10.5m long.
\subsubsection{Python-based Spiral and Linear Simulation}

\begin{figure}[t]
  \centering
  \includegraphics[width=\linewidth,height=0.2\textheight,keepaspectratio]{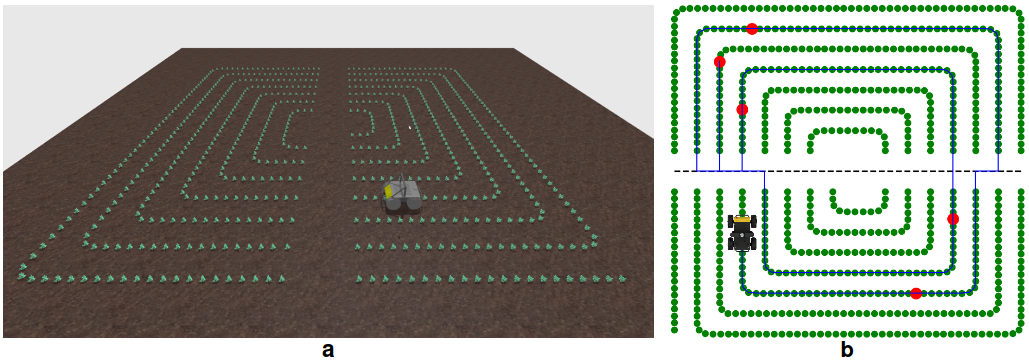}
  \caption{Simulation environments: (a) Spiral crop field in Gazebo simulator with the Husky robot and
           (b) Python-based spiral simulator for large-scale evaluations.}
  \label{fig:exp_envs}
\end{figure}

The Python-based simulator was developed to provide an abstract yet scalable evaluation framework. It reproduced both linear and spiral layouts using exact segment definitions, enabling rapid comparison of coverage, trajectory length, and turning overheads without the computational load of the Gazebo 3D environment. The framework also facilitated the generation of spiral layouts at different scales and the seamless addition of multiple robots, without the time-intensive modelling required in Gazebo. Fig.~\ref{fig:exp_envs}.b illustrates the spiral layout in the Python simulator.

\subsection{Experimental 1: Field Navigation}
\subsubsection{Full Field Navigation}
The first experiment assessed full-field navigation, representing operations such as planting, broad-acre spraying, and autonomous harvesting, where complete coverage and minimal travel are critical. Both the proposed square-spiral layout and the conventional linear field layout were tested on identical land area and row spacing. For the linear field, two end-row manoeuvre strategies were evaluated: wide looping $\Omega$-turns, traditionally used for Ackermann-steering robots, and compact $180^{\circ}$ U-turns designed to reduce distance and time. Under the same field dimensions, the linear layout required fifteen $180^{\circ}$ turns (equivalent in angular effort to thirty $90^{\circ}$ rotations), whereas the \(\mathcal{S}(0.75,11.5)\) square-spiral layout required twenty-eight $90^{\circ}$ turns. This difference slightly reduces the total turning workload in favour of the spiral configuration.

\begin{table}[t]
\centering
\caption{Overall Navigation Performance Across Field Layouts (Identical Land Area and Row Spacing), evaluated with evo Absolute Pose Error metric.}
\label{tab:linear-spiral-comparison}
\resizebox{\columnwidth}{!}{%
\begin{tabular}{lccc}
\hline
\textbf{Layout / Turn Type} & \textbf{Total Distance (m)} & \textbf{Total Time (min)} & \textbf{APE-RMSE (m)} \\
\hline
Linear — $\Omega$-turn & $224.6$ & $18.6$ & $1.65$ \\
Linear — U-turn        & $201.9$ & $15.1$ & $0.43$ \\
Spiral (90$^{\circ}$ turns) & $197.8$ & $14.9$ & $0.38$ \\
\hline
\end{tabular}%
}
\end{table}

\begin{figure}[t]
  \centering
  \includegraphics[width=\linewidth]{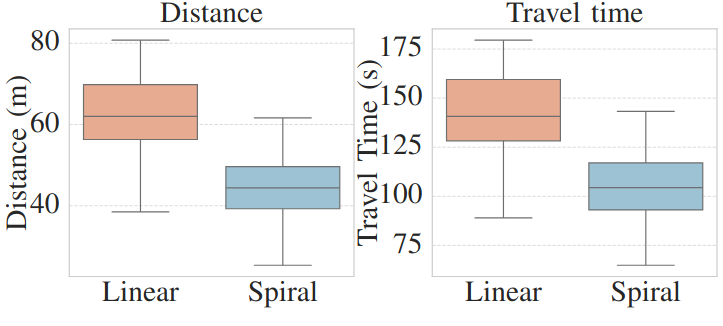}
  \caption{Travel distance and time per batch for waypoint navigation in linear and spiral layouts. Boxplots summarise results from 100 batches generated from an identical set of 500 waypoints.}
  \label{fig:waypoint_boxplots}
\end{figure}

Full-field coverage distance and time highlight the main trade-offs between strategies. The $\Omega$-turn approach incurred greater travel and longer duration due to its wide looping manoeuvres, whereas the U-turn strategy largely eliminated this overhead. The square-spiral layout achieved slightly lower coverage distance and time (197.8 m, 14.9 min) compared to the linear U-turn case (201.9 m, 15.1 min), though the difference was marginal ($\approx$2.0\% in distance and 1.3\% in time). Overall, for full-field coverage tasks, a well-optimised linear field layout with U-turns performs nearly on par with the spiral design, while the spiral remains attractive when broader operational factors are taken into account. Row-following accuracy, assessed using the root mean square error (RMSE) of the \textit{evo} absolute pose error (APE) metric in Table~\ref{tab:linear-spiral-comparison}, was also comparable across layouts, with only a minor increase observed for the linear layout under the $\Omega$-turn strategy.

\subsubsection{Waypoint Navigation}

To evaluate layout-dependent efficiency in precision tasks such as inspection and
targeted spraying, waypoint navigation experiments were performed in both
linear and spiral field configurations. A total of 500 identical waypoints were assigned
for both layouts to ensure direct comparability. These were partitioned into 100 sequential
batches of five targets each, with navigation completed batch by batch. Metrics of interest
were \emph{total travel distance} and \emph{total travel time}, as these directly capture spatial
efficiency and operational duration.

Fig.~\ref{fig:waypoint_boxplots} presents boxplots of batch-wise distance
and travel time. The linear layout exhibited higher medians and broader variance in both
metrics, indicating less predictable performance. In contrast, the spiral
layout demonstrated shorter paths and more stable travel times, underscoring the benefits
of its geometric regularity, especially for targetted tasks that require waypoint navigation.

The mean batch-wise distance was reduced from approximately 62.1\,m in the linear layout
to 44.5\,m in the spiral layout, corresponding to a path efficiency gain of \textbf{28.3\%}.
Travel times showed a similar trend, with an estimated reduction of
\textbf{$\sim$25\%}, highlighting the reduced turning manoeuvres and spatial compactness of square-spiral
layout for robot navigation. These findings confirm that the square-spiral layout offers substantial operational
savings in distance and time during waypoint-based navigation. The results further suggest
that energy consumption is implicitly reduced, since both metrics directly scale with
mechanical workload. Thus, square-spiral layout provides not only faster and more consistent
navigation but also improved sustainability for repeated agricultural tasks.

\subsection{Experimental 2: Crop Row Detection}
We evaluated the proposed DH-ResNet18 for multi-task perception, using its regression head to predict ten $(u,v)$ waypoint pairs and its classification head to identify the scene type (straight segment, left bend, or right bend).  
The predicted waypoints were converted into a central crop-row estimate, and a least-squares line fit was applied to derive two parameters: the angular deviation $\theta$ (relative to the vertical image axis) and the bottom image border intersection coordinate $L_{x}$, defined as the $x$-position where the fitted line meets the lower image boundary.

Following the crop row detection performance metric introduced by  de Silva \etal~\cite{de2024vision}, we computed absolute errors $\Delta\theta$ (degrees) and $\Delta L_{x}$ (pixels) and summarized the crop row detection performance with Equation~\ref{eq:eps_metric}. 

\begin{equation}
\label{eq:eps_metric}
\epsilon = 1 - \frac{1}{2N}\sum_{i=1}^{N}
\left(
\frac{\Delta\theta_i}{\Delta\theta_{\max}} +
\frac{\Delta L_{x,i}}{\Delta L_{x,\max}}
\right),
\end{equation}

where $N$ is the number of images, and $\Delta\theta_{\max}, \Delta L_{x,\max}$ are the dataset-wide maximum errors.

\begin{table}[t]
\centering
\caption{Crop row detection performance comparison on straight segments of spiral layout. $\Delta\theta_{avg}$ in degrees; $\Delta L_{x,avg}$ in pixels; $\epsilon$ expressed as a percentage.}
\label{tab:straight_only}
\setlength{\tabcolsep}{10pt}
\footnotesize
\begin{tabular*}{\columnwidth}{@{\extracolsep{\fill}}lrrr}
\toprule
Method & $\Delta\theta_{avg}$ & $\Delta L_{x,avg}$ & $\epsilon$ (\%) \\
\midrule
Triangle Scan baseline    & 1.74 & 9.49  & 95.10 \\
Proposed DH-ResNet18       & \textbf{0.87} & \textbf{8.26} & \textbf{96.45} \\
\bottomrule
\end{tabular*}
\end{table}

\begin{figure}[t]
  \centering
  \includegraphics[width=\linewidth]{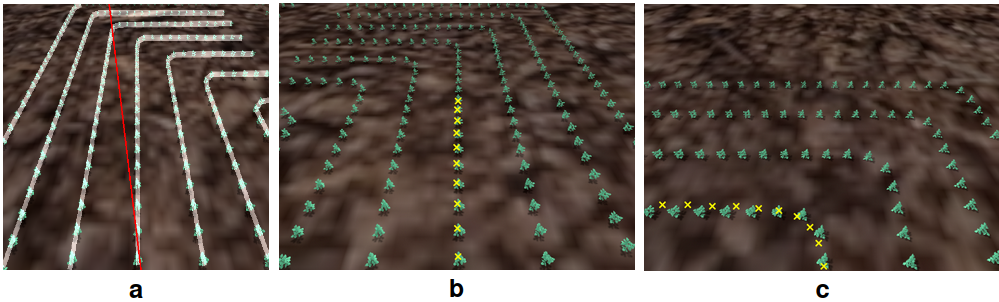}
  \caption{Qualitative comparison of crop-row detection. 
  (a) TSM baseline: fails in spiral layouts because the vertical-peak assumption causes the detected navigation path (red) to deviate from the true navigation path. 
  (b, c) Proposed DH-ResNet18 waypoint regression: robustly predicts the navigation path (yellow waypoints) across both straight and curved crop segments.}
  \label{fig:prediction}
\end{figure}

We collected a dataset of 1,149 images at $480\times640$ resolution from the robot traversing straight segments of the spiral. Images containing curved crop rows were excluded to ensure a fair comparison with the Triangle Scan Method (TSM) baseline. As shown in Table~\ref{tab:straight_only}, the proposed DH-ResNet18 model achieves substantially lower errors than TSM on straight-row navigation. Figure~\ref{fig:prediction}.a illustrates the failure case of TSM: its assumption that crop rows are vertically aligned, as in traditional linear field layouts, leads to erroneous path detection when applied to spiral fields. Specifically, TSM identifies the navigation line by selecting the strongest vertical intensity column from the crop row skeleton mask, an approach that breaks down when rows are horizontal or curved, causing the predicted trajectory (red line) to deviate from the true path. This mode of failure is applicable to any traditional crop row detection pipeline that assume the linear crop row layout. In contrast, Fig.~\ref{fig:prediction}.b and Fig.~\ref{fig:prediction}.c shows the DH-ResNet18 output predicting ego-trajectory waypoints. By directly regressing ten waypoints along the crop line, the network generates a smooth ego-trajectory that remains accurate across both straight and curved segments. This least-squares fitted trajectory yields a 50\% improvement in angular error of navigation path over the baseline. Thus, while the TSM baseline is inherently limited to parallel, linear crop rows, the DH-ResNet18 model generalises effectively to the varying orientations of spiral layouts, making it a more suitable choice for the proposed spiral layout.

\subsection{Experimental 3: Multi-robot task allocation in the square-spiral field layout}
\label{sec:exp_multirobot}

This experiment evaluates the multi-robot task allocation in the square-spiral field layout. We choose a configuration of 3 identical robots in a \(\mathcal{S}(0.75,11.5)\) square-spiral layout performing a waypoint navigation task. Robots were assigned targets in batches using 
either the \emph{Greedy} or the \emph{Hungarian} allocation strategies defined in Section~\ref{sec:multi_robot}. Performance was compared using two indicators: 
the average batch completion time (Eq.~\ref{eq:avg_batch_time}) and the 
workload–fairness coefficient of variation (Eq.~\ref{eq:workload_cv}).

We evaluated the task allocation performance across waypoint batches of varying sizes, using odd counts from 5 to 21 waypoints per batch. For each batch size, multiple independent samples were drawn from a larger waypoint pool. Table~\ref{tab:three_robot} reports the average completion time together with the mean coefficient of variation (CV) of workload fairness, aggregated over all batches at each size.

\begin{table}[t]
\centering
\caption{Navigation and task allocation performance for three robots in the square-spiral field, comparing Greedy and Hungarian methods, averaged across multiple batches of varying sizes.}
\label{tab:three_robot}
\begin{tabular}{ccccc}
\toprule
\multirow{2}{*}{\textbf{Batch Size}} &
\multicolumn{2}{c}{\textbf{Avg.\ Batch Time (s)}} &
\multicolumn{2}{c}{\textbf{Workload Fairness (CV)}} \\
\cmidrule(lr){2-3} \cmidrule(lr){4-5}
& Greedy & Hungarian & Greedy & Hungarian \\
\midrule
5  & 19.96 & 31.72 & 0.339 & 0.468 \\
9  & 29.23 & 43.78 & 0.285 & 0.335 \\
13 & 34.95 & 53.88 & 0.251 & 0.337 \\
17 & 41.01 & 61.97 & 0.242 & 0.300 \\
21 & 42.82 & 66.36 & 0.204 & 0.286 \\
\bottomrule
\end{tabular}
\end{table}

Across all batch sizes, the Greedy allocator consistently achieved batch completion times 33–37\% lower than the Hungarian method, while also maintaining slightly better workload balance (lower mean CV of workload fairness). These findings indicate that, within the square-spiral layout, a three-robot Greedy deployment provides an effective balance between throughput and workload fairness. We present a detailed ablation on the effect of the field size and number of robots on the multi-robot task allocation performance in Section~\ref{sec:abl}.

\subsection{Runtime}

We benchmarked the end-to-end runtime of our navigation pipeline in the \(\mathcal{S}(0.75,11.5)\) square-spiral using Gazebo simulator. The results in \tabref{tab:speed} show an average
execution time per frame of 84--123\,ms, corresponding to 8--12\,Hz frame rate on a laptop-class Intel i7-9750H CPU with a GTX~1660~Ti GPU. Across all trials, the MPC solver dominated the runtime share (60--70\%), the pixel-to-odometry conversion used 25--30\%, and DH-ResNet18 inference accounted for only 8–10\%. These results confirm that the proposed approach is fast enough for real-time online operation on an agricultural autonomous robot, with consistent performance across repeated field traversals.

\begin{table}[t]   
  \caption{Average per-frame runtime across four trials on our test platform
           (Intel i7-9750H CPU @ 2.60\,GHz, NVIDIA GTX 1660 Ti Mobile GPU).
           DH-ResNet18 network was used for waypoint regression and scene classification.}
  \label{tab:speed}
  \setlength{\tabcolsep}{6pt} 
  \footnotesize
  \centering
  \begin{tabular*}{\columnwidth}{@{\extracolsep{\fill}}cccccc}
    \toprule
    \multirow{2}{*}{Trial} & \multicolumn{4}{c}{Time per Component (ms)} & \multirow{2}{*}{FPS} \\
    \cmidrule(lr){2-5}
     & DH\textendash ResNet18 & Pixel$\rightarrow$Odom & MPC & Total &  \\
    \midrule
    1 & 9.13 & 26.97 & 65.51 & 101.62 & 10.17 \\
    2 & 9.28 & 26.84 & 86.42 & 122.54 &  8.16 \\
    3 & 9.05 & 26.08 & 68.09 & 103.22 &  9.69 \\
    4 & 8.86 & 25.08 & 50.79 &  84.72 & 11.80 \\
    \midrule
    Mean & \textbf{9.08} & \textbf{26.24} & \textbf{67.95} & \textbf{103.03} & \textbf{9.96} \\
    \bottomrule
  \end{tabular*}
\end{table}

\section{Ablation Study}
\label{sec:abl}
\subsection{Ablation 1: Scalability with Increasing Spiral Field Size}
To examine scalability, the square-spiral layout was evaluated with increasing spiral sizes of
7, 9, 13, 17, and 21 concentric spiral loops, using the same batch-based waypoint protocol of 500 targets.
The objective was to test whether the efficiency advantages observed in smaller fields
persist in larger deployments.

\begin{figure}[t]
  \centering
  \includegraphics[width=\linewidth]{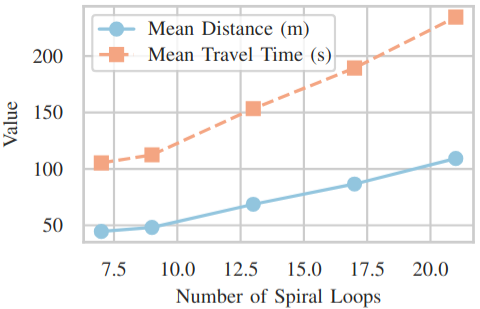}
  \caption{Waypoint navigation performance vs, spiral size (loop count).
           Both mean travel distance and time increase
           approximately linearly as the number of loops grows.}
  \label{fig:ablation_scaling}
\end{figure}

Fig.~\ref{fig:ablation_scaling} summarize the
results. Both travel distance and time increase linearly with loop count. This confirms that spiral navigation scales predictably with field size while maintaining its relative efficiency and consistency.

\subsection{Ablation 2: Effect of Robot Count and Batch Size}
\label{sec:ablation_robotcount}

This ablation evaluates the \emph{Greedy} multi-robot allocator described in
Section~\ref{sec:multi_robot}, investigating how robot count and batch size of targets influence navigation efficiency and workload balance. Robot teams of
\(m \in \{2,3,4\}\) operated in the square–spiral field while processing waypoint batches of sizes between 5–21.

\subsubsection{Average Batch Completion Time}
\begin{figure}[t]
  \centering
  \includegraphics[width=\linewidth]{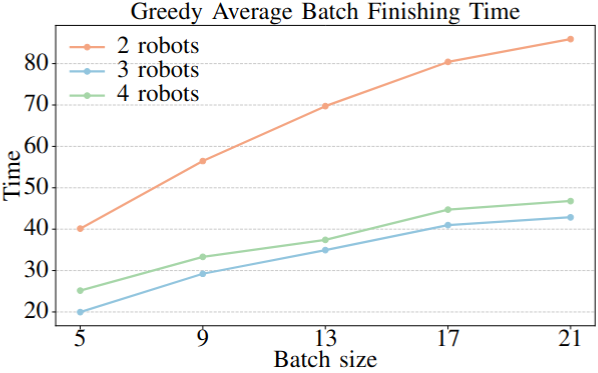}
  \caption{Mean batch completion time (seconds) vs. batch size for different robot team sizes using the Greedy allocator.}
  \label{fig:ablation_time}
\end{figure}

Figure~\ref{fig:ablation_time} shows that increasing the number of robots consistently reduces the mean completion time across all batch sizes. The largest improvement is observed when scaling from two to three robots; for instance, at 21 targets the average time decreases from 86\,s with two robots to 43\,s with three. Introducing a fourth robot yields only marginal gains and, for smaller batch sizes, can even increase completion time, suggesting occasional interference once the workspace becomes saturated.

\subsubsection{Workload Fairness}
\begin{figure}[t]
  \centering
  \includegraphics[width=\linewidth]{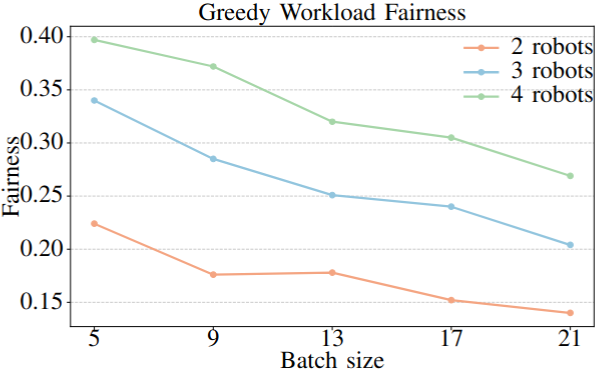}
  \caption{Per-robot done fairness measured as the coefficient of variation (CV) vs. batch size for different robot team sizes. Lower CV indicates more balanced sharing.}
  \label{fig:ablation_cv}
\end{figure}

Figure~\ref{fig:ablation_cv} reports the per-robot workload fairness CV across varying batch sizes with different robot counts. In the workload fairness CV metric, the lower values indicate a more even division of work. Two-robot teams achieve the most balanced allocation, three-robot teams remain reasonably fair, while four-robot teams shows the highest dispersion up to 0.4, reflecting periods of idle time when targets are scarce.

\section{Conclusion}
\label{sec:conclusion}

We introduced a square-spiral crop layout that rethinks field geometry for autonomous agriculture and presented a navigation stack tailored to it. By replacing tractor-oriented headlands and parallel rows with a continuous spiral plus a central tramline, the layout reduces tight turning, shortens transit between targets, and injects geometric variability that mitigates perceptual aliasing. Coupled with our DH-ResNet18 waypoint regressor, pixel-to-odometry projection, A* path planner, and MPC tracker, the system improved trajectory accuracy and reliable real-time performance.

Across identical land areas and row spacing, the square-spiral layout matched or slightly outperformed the linear layout for coverage tasks, and delivered gains for waypoint-based operations: mean travel distance decreased by $\sim$28\% and travel time by $\sim$25\%, with corresponding reductions in mechanical workload (and thus energy use). Multi-robot experiments further demonstrated that rule-constrained planning on the spiral graph supports efficient, collision-aware coordination, with a simple greedy allocator providing strong throughput and reasonable workload balance.

More broadly, our results suggest that redesigning field structure for robots rather than adapting robots to legacy tractor layouts can unlock practical efficiency and robustness benefits. Future work will investigate real field deployments, extend the 2D-to-1D spiral projection for lightweight localization and teach-and-repeat methods, and explore hybrid layouts and crop management practices that combine spiral geometry with agronomic constraints.



\bibliographystyle{plain_abbrv}

\bibliography{new}

\end{document}